%% file: Main_paper.tex
\documentclass[sigconf, 10pt, anon, nonacm]{acmart}
\acmConference[DroNet 2025]{11th Workshop on Micro Aerial Vehicle Networks, Systems, and Applications}{June 2025}{Anaheim, California, USA}
\usepackage{graphicx}
\usepackage{subcaption}
\usepackage{algorithm,algorithmic}
\usepackage{color}
\usepackage{xcolor}
\usepackage{multirow}
\usepackage{caption}
\usepackage[colorinlistoftodos]{todonotes}
\usepackage[normalem]{ulem}

\copyrightyear{2024}
\acmYear{2024}\copyrightyear{2024}
\setcopyright{rightsretained}
\acmConference[DroNet' 24]{The 10th Workshop on Micro Aerial Vehicle Networks, Systems, and Applications}{June 3--7, 2024}{Minato-ku, Tokyo, Japan}
\acmBooktitle{The 10th Workshop on Micro Aerial Vehicle Networks, Systems, and Applications (DroNet' 24), June 3--7, 2024, Minato-ku, Tokyo, Japan}
\acmDOI{10.1145/3661810.3663468}
\acmISBN{979-8-4007-0656-1/24/06}

\begin{document}
\settopmatter{printfolios=true}

\title{Continuous Marine Tracking via Autonomous UAV Handoff}

\author{Heegyeong Kim$^1$, Alice James$^1$, Avishkar Seth$^1$, Endrowednes Kuantama$^1$, Jane Williamson$^2$, Yimeng Feng$^1$, Richard Han$^1$}

\affiliation{%
  \institution{School of Computing$^1$, School of Natural Sciences$^2$}
  \country{Macquarie University, Australia}\\
   }

\renewcommand{\shortauthors}{Kim et al.}

\begin{abstract}

This paper introduces an autonomous UAV vision system for continuous, real-time tracking of marine animals, specifically sharks, in dynamic marine environments. The system integrates an onboard computer with a stabilised RGB-D camera and a custom-trained OSTrack pipeline, enabling visual identification under challenging lighting, occlusion, and sea-state conditions. A key innovation is the inter-UAV handoff protocol, which enables seamless transfer of tracking responsibilities between drones, extending operational coverage beyond single-drone battery limitations. Performance is evaluated on a curated shark dataset of 5,200 frames, achieving a tracking success rate of 81.9\% during real-time flight control at 100 Hz, and robustness to occlusion, illumination variation, and background clutter. We present a seamless UAV handoff framework, where target transfer is attempted via high-confidence feature matching, achieving 82.9\% target coverage. These results confirm the viability of coordinated UAV operations for extended marine tracking and lay the groundwork for scalable, autonomous monitoring.

   
\end{abstract}



\maketitle

\input{intro}
\input{related}

\input{tracking-algo}

\input{system}

\input{results}

\input{conclusion}

\vspace{-2mm}
\bibliographystyle{splncs04}
\bibliography{refs_dronet}

\end{document}

%% file: intro.tex
\section{INTRODUCTION}

\begin{figure}
    \centering
    \includegraphics[width=0.9\linewidth]{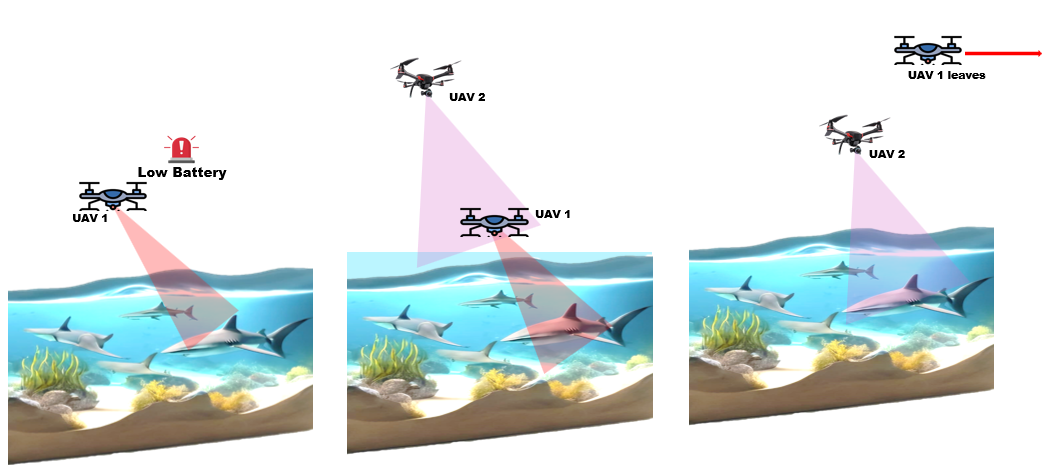}
    \caption{The principle of Continuous data handoff for marine monitoring}
    \label{overall}
\vspace{-5mm}
\end{figure}

Effective monitoring of marine animals, such as sharks, is essential for conserving marine biodiversity and maintaining healthy marine ecosystems~\cite{biodiversitymarine,williamson2024environmental}. Sharks display intricate and dynamic movement patterns, which present significant challenges for ecological research and public safety management~\cite{elliott2022satellite,jorgensen2022emergent}. The demand for precise and real-time monitoring is especially critical in Australia, a region that experiences a high frequency of human-shark interactions. Over the past decade, Australia has recorded an average of approximately 20 shark-related human injury incidents annually~\cite{TarongaAustralianSharkIncidentDatabase}, highlighting the urgent need for enhanced monitoring strategies. Traditional tracking methods, including manual observation and acoustic telemetry, are often limited by human error, spatial constraints, and high operational costs, leading to potentially incomplete or delayed data~\cite{villon2024toward}. To address these limitations, the integration of Uncrewed Aerial Vehicles (UAVs), specifically Vertical Take-Off and Landing (VTOL) platforms like quadcopters and hexacopters, presents a promising solution~\cite{butcher2021drone}. These vehicles are capable of offering extended flight endurance, precise hover capabilities, and high-resolution visual data acquisition, surpassing the limitations of satellite tracking and other traditional methods. UAV-based monitoring systems enable real-time, accurate, and cost-efficient observation of shark populations by combining high-resolution imaging with robust tracking capabilities. This approach enhances coastal safety by supporting timely decision-making and improves ecological research through the provision of high-resolution spatial and temporal data on shark behavior~\cite{lalgudi2025zero}. By addressing the limitations of manual observation, it offers a more efficient, scalable, and reliable framework for understanding shark movement patterns and habitat use.


Maintaining continuous and accurate real-time marine tracking presents a significant challenge. Oceanic environments introduce inherent complexities, including water occlusions, submerged obstacles, and potential errors in species and individual identification, all of which can compromise data integrity~\cite{varini2024sharktrack}. Furthermore, the vast spatial scales requiring surveillance pose a substantial scalability hurdle. Both aerial and underwater platforms face environmental limitations, such as variable water clarity, adverse weather, and signal interference, hindering effective tracking~\cite{ganie2025unmanned}.

Current tracking solutions typically rely on a single UAV, but these are constrained by limited flight times less than an hour due to the battery. This restricts the duration of continuous monitoring. To address this, we propose a cooperative multi UAV monitoring solution where a second UAV takes over tracking duties when a first UAV has a depleted battery. The critical challenge lies in achieving seamless handoff target tracking, ensuring that the second UAV identifies and follows the same target as the first UAV. As illustrated in Fig ~\ref{overall}, we introduce the vision based handover protocol that leverages real time feature matching and inter-drone coordination to maintain continuous tracking across agents. 



The key contributions of this study include:
\begin{itemize}
    \item \textbf{Seamless Inter-UAV Handover for Extended Tracking:} This paper proposes a synchronised handover protocol enabling continuous shark tracking across UAVs.
    \item \textbf{Embedded UAV Vision System for Real-Time Marine Tracking:} A stabilised RGB-D camera with an onboard computer enables robust, low-altitude shark tracking in dynamic motion.
    \item \textbf{Shark Tracking Dataset:} A custom-annotated aerial dataset with challenging environmental occlusion, motion blur, and coastal variability, used to evaulate real-time tracking performance in a realistic marine setting.
    \end{itemize} 

%% file: related.tex
\vspace{-3mm}
\section{RELATED WORK}
Previous studies on tracking animals have utilized various methodologies, employing computer vision techniques and deep learning models for UAV-based detection and classification of wildlife in their natural habitats, including koalas, pigeons, and fish~\cite{liu2024deep,winsen2022automated}. Specifically within marine environments, UAV-based tracking offers an effective platform for deploying deep learning techniques. Among these, the YOLO model, utilising a convolutional neural network (CNN) backbone, has emerged as a popular choice. By framing object detection as a single-stage regression problem, YOLO enables real-time identification and localisation of marine species with high computational efficiency
~\cite{lalgudi2025zero}. For example,~\cite{dujon2021machine} presents a lightweight CNN architecture for processing UAV-derived imagery to detect and monitor the behaviour of multiple marine taxa. The major challenge identified was the varying detection accuracy of the CNN, which significantly depended on species morphology, behaviour, spacing, and habitat complexity, thereby limiting the model’s ability to generalise across diverse marine environments. Similarly,~\cite{purcell2022assessing} demonstrated a deep learning method, RetinaNet with ResNet-50 and MobileNet V1, can achieve real-time shark species identification in drone video, but emphasised the critical need for carefully curated training data and environmental tuning to ensure consistent detection accuracy in varying field conditions.

Similarly, authors in~\cite{sharma2021sharkspotter} developed an AI-based shark detection solution with reportedly a 90\% accuracy. While the practical achievements are noteworthy for an early-stage solution, the system required refinement to ensure reliability in diverse environments. It proved unreliable when deployed to new locations, and the computing hardware was not portable or robust enough to withstand the harsh beach conditions. Furthermore, the SharkEye platform~\cite{gorkin2020sharkeye} utilised blimp-mounted cameras and a YOLO-based deep neural network to detect sharks and other marine life, validating human observation as a benchmark and developing a custom algorithm trained on manually annotated video data. However, minimal training footage and environmental constraints limited the system's reliability, highlighting the need for more extensive data and robust deployment conditions.

Another study~\cite{colefax2023utility} employed spectral filtering and machine learning techniques to assess the reliability of drone-based marine fauna detection, revealing that RGB sensors without spectral restrictions yielded the best performance. However, a key limitation of the study was that spectral filtering and polarising treatments decreased detection accuracy, suggesting no significant benefit over standard RGB cameras in the given environment. Current literature identifies key challenges in shark tracking using computer vision, including algorithm selection, weather resilience, robust training, continuous tracking, and real-time hardware constraints. While many drone-based detection systems exist, few ensure reliable, uninterrupted tracking, largely due to limited flight times that disrupt long-term monitoring.


%% file: tracking-algo.tex
\vspace{-3mm}
\section{TRACKING ALGORITHM}
In this study, we evaluate five state-of-the-art visual tracking algorithms —OSTrack, AiATrack, Track-Anything, MixFormer, and SiamAPN —to cover a range of architecture paradigms and task formulations relevant to drone-based tracking in natural environments.

\vspace{-3mm}
\subsection{Tracking Method Comparison}


\textbf{OSTrack} ~\cite{ye2022jointfeaturelearningrelation} is an end-to-end transformer-based single object tracker applying unified attention-based model. It learns tracking behavior directly from the data without depending on handcrafted stages. As focusing on key visual elements in each frame, it can mitigate distractions from water texture and motion. 
 \textbf{MixFormer} ~\cite{mixformer} is a transformer model with early feature fusion integrating template and search features early in the tracking pipeline which improves both temporal stability and robustness to abrupt target appearance changes. This is beneficial when the target suddenly changes its pose, orientation or visibility. 

\textbf{SiamAPN}~\cite{cao2021siamapn} is designed for real-time tracking on aerial platforms, making it well-suited for onboard drone deployment. Its fast processing and low computational demand support high frame-rate tracking, which is crucial for capturing fast, erratic animal motion. Its attention-based aggregation improves robustness in visually unstable marine environments with shifting light and wave patterns.

\textbf{AiATrack}~\cite{gao2022aiatrack} enhances the standard attention mechanism through nested attention modules, enabling more effective separation of target and background in visually complex or cluttered scenes. This design is particularly suited to marine tracking scenarios involving camouflage, partial submersion, or visual blending with water textures. \textbf{Track-Anything} ~\cite{yang2023track}  enables flexible, interactive tracking through segmentation prompts, making it valuable when the target is partially occluded or temporarily lost. Its segmentation improves accuracy in low-contrast underwater scenes, and its re-identification capability helps recover tracking after brief target disappearance.
\vspace{-2mm}
\subsection{Dataset}

The dataset used in this study consists of real-world aerial footage capturing a shark swimming in shallow coastal waters, recorded from an overhead perspective. The video was originally collected by marine biologists for ecological research and later repurposed for the evaluation of visual tracking algorithms. This footage provides a realistic marine setting that closely resembles drone-based tracking conditions.

The video was recorded under sunny conditions, featuring visible sun glare, surface reflections, and wave motion across a shallow seabed populated with rocks and coral formations. These natural variations create significant visual complexity, including partial occlusion, low contrast between the target and background, and dynamic visual noise caused by water movement. Such characteristics make the dataset well-suited for testing tracking robustness in real-world scenarios.

Each frame is annotated with shark bounding boxes and frame-to-frame ID consistency, enabling evaluation of both detection accuracy and temporal tracking. the dataset includes 6928 frames, corresponding to a 2-minute and 25-second video sequence, resulting in an approximate frame rate of 50 fps. As these include unpredictable target motion, brief occlusions due to underwater terrain, and visual blending with background textures, the dataset provides a valuable benchmark for assessing the performance of tracking algorithms in marine environments from an aerial perspective.
\vspace{-5mm}
\subsection{Matching Algorithm}

To enable efficient feature matching between frames captured from different drones, we adopted the \textbf{ORB (Oriented FAST and Rotated BRIEF)} algorithm. ORB is a fast and lightweight feature extraction and matching method that combines the FAST keypoint detector with the BRIEF descriptor, while adding rotation and scale invariance ~\cite{rublee2011orb}. ORB was designed as a computationally efficient alternative to more complex algorithms such as SIFT and SURF. Its suitability for real-time applications and low-power environments, such as onboard drone systems, makes it an ideal choice for our system. ORB provides competitive accuracy compared to SIFT, particularly in scenarios involving moderate rotation and noise, while offering significantly better performance in terms of speed ~\cite{karami2017image}.

%% file: system.tex
\section{cooperative continuous tracking}
\begin{figure}[!htpb]
    \centering
    \includegraphics[width=\linewidth]{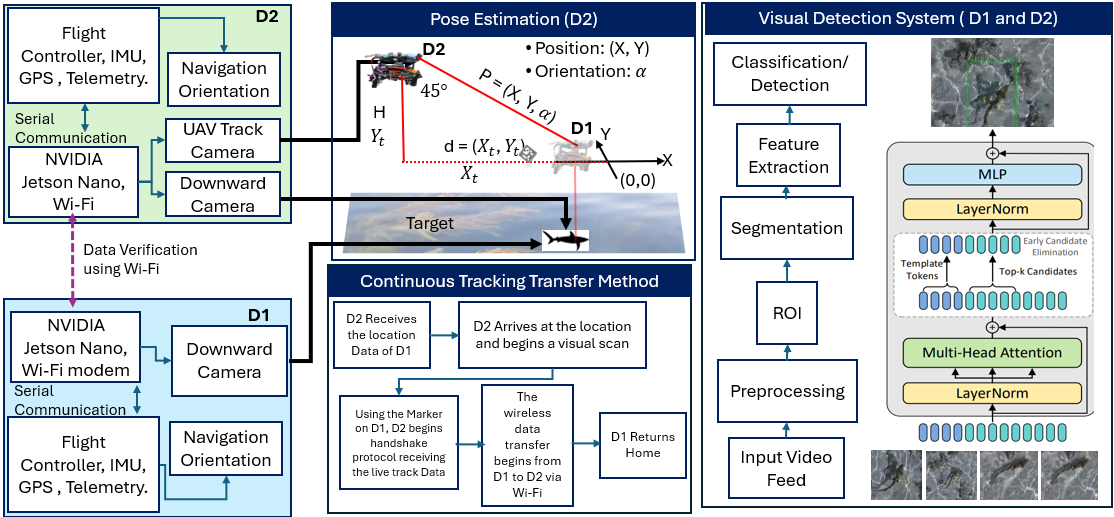}
    \caption{Proposed system architecture of the continuous tracking and transfer method}
    \label{system}
\end{figure}

The proposed system comprises two quadrotor UAVs, D1 and D2, each with a 425 mm frame, equipped for autonomous marine monitoring. Each UAV integrates a Pixhawk flight controller with IMU and GPS modules to handle low-level motor control and orientation stabilisation. An onboard NVIDIA Jetson Nano (8 GB) enables real-time visual processing, navigation, and decision-making. A 45$^\circ$-tilted RGB-D camera mounted on each UAV provides depth-augmented imagery optimised for marine surface tracking. Communication is managed via a Wi-Fi module, allowing inter-UAV coordination and uplink to a ground control station. MAVROS (via ROS Noetic) facilitates data exchange between the Jetson and the Pixhawk at a 115200 baud rate, supporting navigation, sensing, and mission updates. The overall system workflow is illustrated in Fig.~\ref{system}.

\vspace{-2mm}
Each UAV runs a ROS-based tracking pipeline in the software architecture, combining target detection and state estimation for closed-loop control. Visual data from the RGB-D camera is processed onboard using deep tracking algorithms, including OSTrack, MixFormer, AiATrack, and XMem variants. These models are containerised and optimised for Jetson's GPU resources. The UAV navigates autonomously based on visual detections, updating real-time target position estimates. Each UAV independently runs the most effective tracking algorithm, sharing target state estimates via inter-UAV communication. Handover is achieved by matching the high-confidence visual features of the tracked shark, enabling seamless transition and uninterrupted monitoring beyond the endurance limits of a single UAV.

In the pose estimation process, Drone D2 tracks a shark located at the origin \((0, 0)\) on a 2D coordinate plane, continuously adjusting its position \((X, Y)\) to keep the shark centered in its field of view in a 480x640 resolution of frame. An ArUco marker is placed on top of D2, aiding in D1's localization. This marker, along with GPS data, allows Drone D1 to determine D2’s position and orientation, denoted by \(\alpha\). D1 uses its camera to estimate the pose of D2 as \(P = (X, Y, \alpha)\) and remains aligned parallel to D2, maintaining a horizontal offset \(X_t\) and a vertical offset \(Y_t\) (altitude difference). The camera on D1 is positioned at a \(45^\circ\) angle, providing an optimal view of both D2 and the target tracking object.

\subsection{Handoff Protocol}
The Continuous Tracking Transfer Method enables a smooth transition of tracking responsibilities between the drones during the mission. First, Drone D2 receives real-time location data from D1 and navigates to its position. Once D2 arrives, it uses onboard vision to detect the ArUco marker on D1. Following marker detection, a handshake protocol is initiated to synchronize the drones. D1 then wirelessly transmits live tracking data to D2 via Wi-Fi, enabling D2 to assume the tracking task.

\begin{figure}[!htpb]
    \centering
    \includegraphics[width=0.8\linewidth]{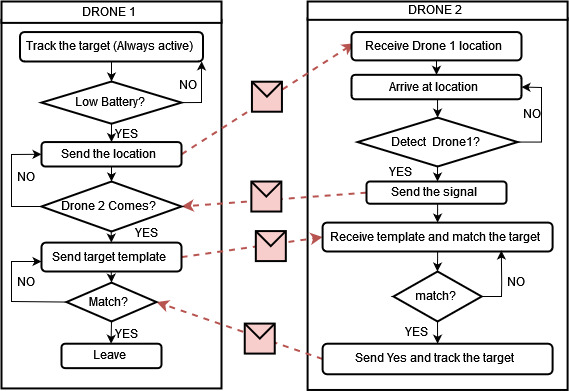}
    \caption{Seamless Data Transfer Flowchart}
    \label{fig:enter-label}
    \vspace{-2mm}
\end{figure}

With D2 now handling the tracking, Drone D1 can safely return to its base station. This transfer method ensures continuous target monitoring, allowing the drones to alternate tracking duties - meaning that D2 is now referred to as D1. By distributing the workload, each drone conserves battery life and extends mission duration, resulting in a more efficient and effective monitoring system.



%% file: results.tex
\section{RESULTS AND VALIDATION}
\subsection{Tracking Performance}
This section presents the evaluation of five state-of-art trackers-\textbf{OSTrack, MixFormer, SiamAPN, AiATrack}, and \textbf{Track Anything}- on our custom dataset composed of 5,200 frames 
of aerial marine footage. The dataset captures a shark in natural conditions, exhibiting real-world challenges such as partial occlusion, submersion, reflection, and complex background textures. These conditions enable us to evaluate the practical applicability of each tracker in realistic environmental monitoring tasks using UAVS. We evaluate performance across two dimensions: \textbf{1. Tracking Accuracy:} How well each model localized the target spatially and temporally. \textbf{2. Robustness:} How each model handled long term tracking, occlusion recovery, and target disappearance. 

%
\begin{table}[h]
\vspace{-1mm}
\caption{Accuracy of trackers}
\resizebox{0.9\columnwidth}{!}{%
\begin{tabular}{|l|l|l|l|l|l|}
\hline
Tracker &
  IoU(avg) &
  \begin{tabular}[c]{@{}l@{}}Success\\  Rate\end{tabular} &
  \begin{tabular}[c]{@{}l@{}}Precision\\ @20px\end{tabular} &
  GT Cover\\ \hline
OSTrack        & 0.612  & 74.8\%  & 57.2\% & \textbf{85.7\%} \\ \hline
MixFormer      & \textbf{0.655} & \textbf{81.9\%} & \textbf{59.5\%} & 78.9 \%         \\ \hline
SiamAPN        & 0.044    & 2.1\%   & 3.4\%   & 4.5\%   \\ \hline
AiATrack       & 0.079    & 2.4\%   & 0.7\%   & 20.8\% \\ \hline
Track Anything & 0.242    & 29.0\%   & 27.0\%   & 26.9\%  \\ \hline
\end{tabular}%
}
\label{accuracy}
\end{table}
\vspace{-1mm}

\textbf{Tracking Accuracy.} We evaluated the tracker's spatial alignment and prediction accuracy using four metrics: Mean Intersection-over-Union (IoU), Success Rate (defined as percentage of frames with an IoU higher than 0.5), and Precision at 20 pixels as depicted in Table ~\ref{accuracy}. To complement these conventional measures, we also introduced a Ground Truth Coverage metric (GT Cover \%), which quantifies how often and how fully the predicted bounding box encompasses the ground truth object, regardless of its exact shape or scale.
 
 Among all models, \textbf{MixFormer} has the highest average IoU of 0.655, success rate of $81.9 \%$ and produced the most precise results of $59.5 \%$ - indicating superior accuracy in consistently localizing the target. \textbf{OSTrack} was close behind with IoU of 0.612 but outperformed others in average ground truth coverage $(85.7\%)$. On the other hand, SiamAPN, AiATrack, and Track-Anything underperformed with higher failure, which has IoU of 0, and low success rate, indicating instability in dynamic scenes.

\begin{table}[h]
\caption{Robustness of trackers}
\resizebox{\columnwidth}{!}{%
\begin{tabular}{|l|c|c|c|}
\hline
Tracker &
  \begin{tabular}[c]{@{}c@{}}Failure Count\\ (IoU \textless{}0.1)~[frames]\end{tabular} &
  \begin{tabular}[c]{@{}c@{}}Longest Failure Streak\\ (IoU \textless{}0.1)~[frames]\end{tabular} &
\begin{tabular}[c]{@{}c@{}}Avg Drift\\ ~[px] \end{tabular} \\ \hline
OSTrack        & \textbf{467} & \textbf{251} & \textbf{5.6} \\ \hline
MixFormer      & 512          & 511          & 6.2          \\ \hline
SiamAPN        & 4505         & 4339         & 2.8          \\ \hline
AiATrack       & 2853         & 655          & 2.2          \\ \hline
Track Anything & 3630         & 3439         & 809.1        \\ \hline
\end{tabular}%
}
\label{Robustness}
\end{table}

 \textbf{Robustness.} We evaluated three complementary metrics: Failure count, total number of frames where the tracker completely failed to align with the target, Longest failure streak measuring the worst-case duration of continuous tracking failure, and Average drift @20px which shows the average center distance between prediction and ground truth when predictions are within a 20-pixel radius. This factor reflects how tightly the tracker stays on target during near correct frames. 
 
 As seen in Table~\ref{Robustness}, \textbf{OSTrack} exhibited the most consistent robustness, achieving the lowest number of total failures with 467 frames and the shortest maximum failure streak with 251 frames. This indicates that OSTrack can quickly recover from tracking interruptions and maintain stable performance, even when the target becomes partially occluded or visually ambiguous. In contrast, \textbf{MixFormer}, despite performing well in accuracy metrics, exhibited a failure streak of 511 frames, nearly double that of OSTrack suggesting MixFormer is more prone to prolonged tracking loss, which undermines its reliability during real-world deployments. 
\vspace{-1mm}
 \begin{figure}[!htpb]
    \centering
    \includegraphics[width=0.8\linewidth]{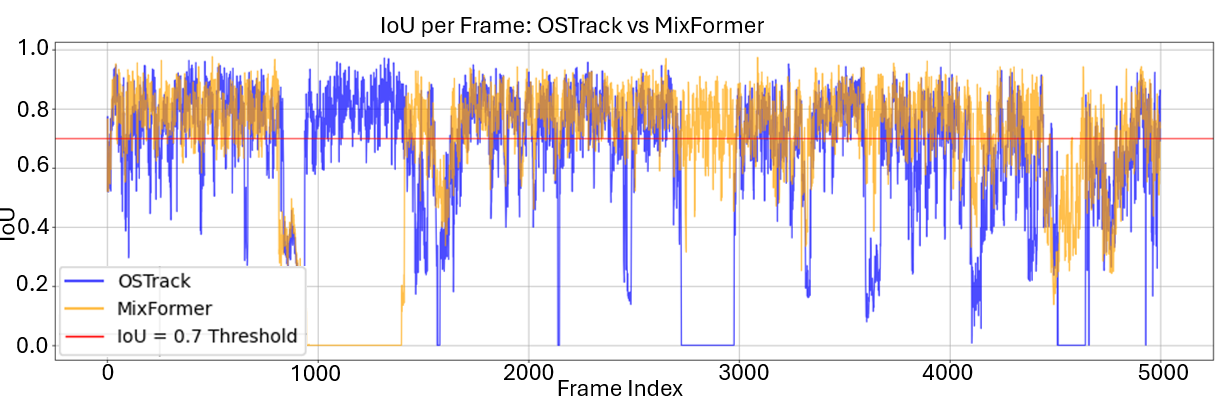}
    \includegraphics[width=0.8\linewidth]{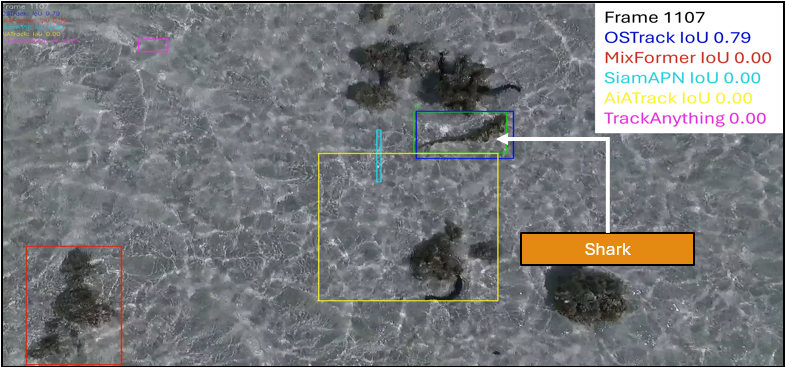}
    \caption{OSTracker performance}
    \label{OSonly}
\end{figure}
\vspace{-3mm}
In terms of drift, OSTrack maintained a low drift of 5.6 pixels, confirming its spatial consistency when tracking correctly. SiamAPN and AiATrack have very low drift values, 2.8 and 2.2 pixels, respectively. These values are considered unreliable indicators of robustness due to both models exhibiting extremely high failure counts and low success rates, which are calculated only on a minimal subset of successful frames. As emphasized in prior tracking literature, low error without frequency of success leads to a misleading conclusion ~\cite{Kristan2018a}. 

Fig.~\ref{OSonly} illustrates key observations supporting that OSTrack is the most reliable among the tested methods. The upper panel compares the IoU of OSTrack and MixFormer over time. While both trackers achieve comparable average accuracy, MixFormer lost the target for longer time. The lower panel highlights that only OSTrack successfully identifies the correct target while all other trackers are mislead by visually similar background element. These results demonstrates that OSTrack has the strongest balance between recovery speed, error consistency, and false positive suppression, making it the most reliable choice for long term tracking in dynamic marine conditions. 

\vspace{-2mm}
\subsection{UAV Data Handover}
\begin{figure}[!htpb]
\vspace{-2mm}
    \centering
    \includegraphics[width=\linewidth]{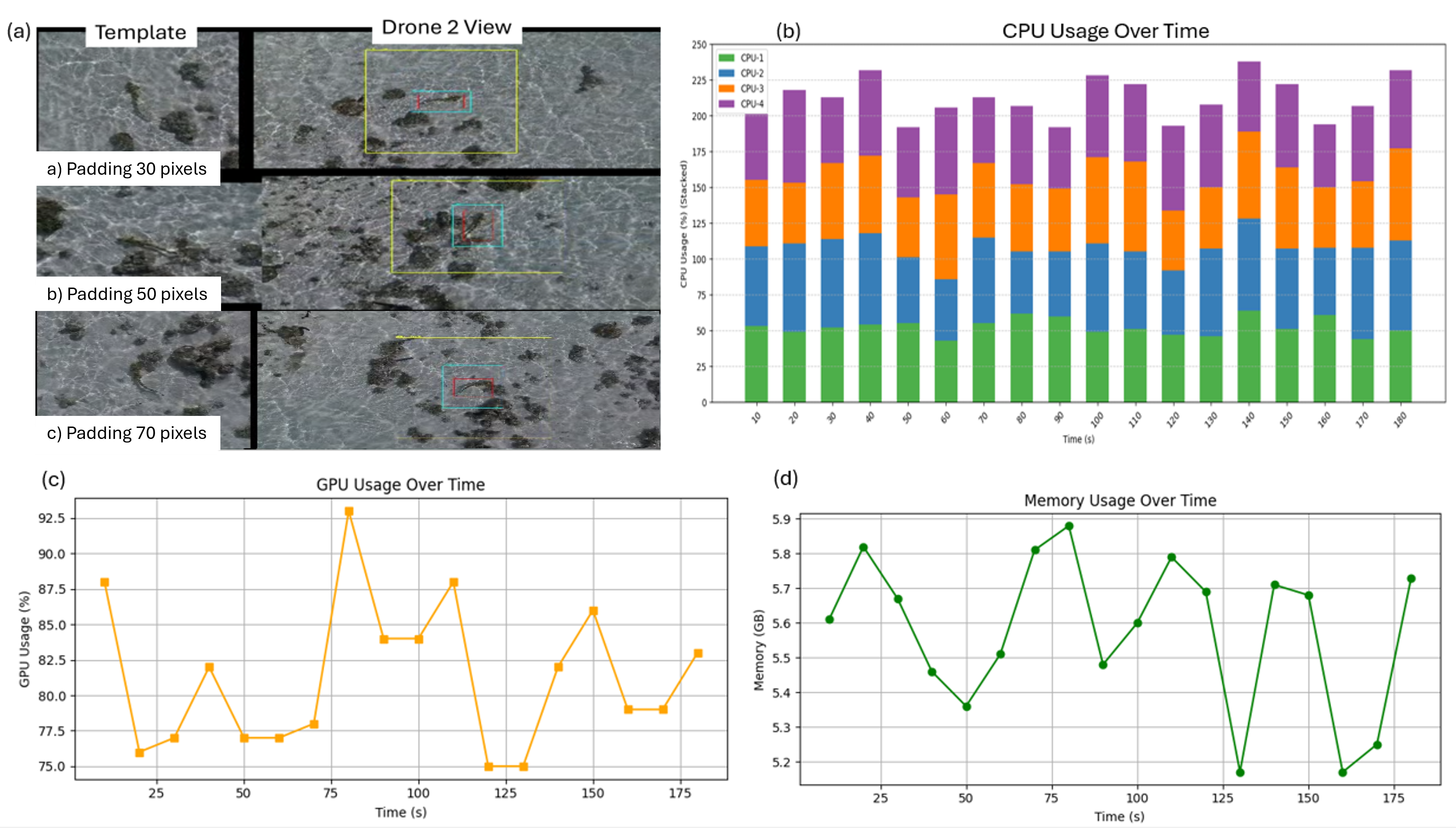}
    \caption{Handover: (a) Feature matching; System performance: (b) CPU (c) Memory (d) GPU}
    \label{Matching+perf}
    \vspace{-1mm}
\end{figure}

Selecting OSTrack, we next adopted a two-stage extraction strategy, where we first cropped a broad region of interest using fixed padding of 300 pixels and then attempted to localize the shark within this region using tighter templates with varying paddings, 30, 50, and 100 pixels. As illustrated in Fig.~\ref{Matching+perf} (a), the left images show the template extracted from D1, the right images represent the view from D2. The yellow bounding box indicates the cropped region of interest, the blue box indicates the matched prediction, and the red box represents the ground truth annotation. 

As shown in Table ~\ref{matching}, larger template padding led to slightly improved performance. With a 70-pixel template, the system achieved the highest target coverage of 82.98\%, with the longest matching streak of 236 frames. Given the different viewing angles between D1 and D2, the matched bounding box consistently included the target, even when tight alignment was not achieved.


\begin{table}[h]
\caption{Matching performance}
\resizebox{0.7\columnwidth}{!}{%
\begin{tabular}{|c|c|c|}
\hline
\# Padding [pixels] & Target Cover  & \begin{tabular}[c]{@{}c@{}}Longest Success Streak \\ ~[frames] \end{tabular} \\ \hline
30             & 68.37\% & 94 \\ \hline
50             & 80.21\%   & 92  \\ \hline
70              & 82.98\%  &  236 \\ \hline

\end{tabular}%
}
\label{matching}
\end{table}
\vspace{-3mm}


The onboard system tests were performed in a hardware-in-the-loop (HIL) environment, which included visual tracking, feature matching, inter-UAV coordination, and ROS-based navigation logic, resulting in stable end-to-end performance at 4–5 Hz, even under continuous coastal operation. This rate reflects the complete processing loop from image capture to trajectory adjustment, driven by OSTrack with the '$vitb\_256\_mae\_ce\_32x4\_ep100$' configuration. While the high-level visual tracking pipeline operates at this frequency, low-level control loops for attitude and position—handled by the Ardupilot flight stack—ran independently at 100+ Hz, ensuring responsive and stable flight behaviour. 

The system performance for 180 seconds is shown in Fig.~\ref{Matching+perf} (b-d). The full-load operation, which included visual tracking, ROS flight node execution, and MAVROS communication, resulted in an average power consumption of 7.4 W. The peak draw reached 8.9 W. GPU utilisation remained consistently high, ranging from 75 to 90\%. At the same time, memory usage stabilised at around 5.5 GB, indicating efficient memory management under sustained inference. The CPU managed orchestration and feature matching with a moderate load, averaging below 60\%. These metrics demonstrate a well-balanced trade-off between computational throughput and energy efficiency for real-time tracking and coordination in compact UAV systems under operational constraints.

%% file: conclusion.tex
\vspace{-3mm}
\section{CONCLUSION}
This study introduced a UAV-based vision framework for autonomous marine animal tracking, achieving 74.8\% success rate and real-time performance in dynamic marine environments. The system leveraged OSTrack, exhibiting glare, occlusion, and motion drift robustness. A UAV-to-UAV handover strategy was evaluated using a two stage template matching approach, where the matched bounding box achieved over 82.9\% target coverage maintaining a maximum continuous match of 236 frames, equivalent to more than 7 seconds. Onboard deployment was realised using the Jetson Nano platform, maintaining effective tracking accuracy after optimisation, with a minimal reduction in detection performance. These outcomes validate the system's practical feasibility and highlight its novelty in enabling lightweight, scalable, and sustained autonomous monitoring of marine ecosystems using coordinated UAVs.